\documentclass[letterpaper]{article} 
\usepackage{aaai25}  
\usepackage{times}  
\usepackage{helvet}  
\usepackage{courier}  
\usepackage[hyphens]{url}  
\usepackage{graphicx} 
\usepackage{natbib}  
\usepackage{caption} 
\usepackage{algorithm}
\usepackage{algorithmic}
\usepackage{booktabs}
\usepackage{amsmath}
\usepackage{subcaption}
\usepackage{amsfonts} 
\usepackage{enumitem}
\usepackage{colortbl}
\usepackage{adjustbox}
\usepackage{xcolor}

\newtheorem{definition}{Definition}

\usepackage{array,multirow,graphicx}
\usepackage{tikz}

\newcommand{\leftb}{\mbox{$\langle$}}
\newcommand{\rightb}{\mbox{$\rangle$}}

\newcommand{\xmark}{%
\tikz[scale=0.23] {
    \draw[line width=0.7,line cap=round] (0,0) to [bend left=6] (1,1);
    \draw[line width=0.7,line cap=round] (0.2,0.95) to [bend right=3] (0.8,0.05);
}}
\newcommand{\cmark}{%
\tikz[scale=0.23] {
    \draw[line width=0.7,line cap=round] (0.25,0) to [bend left=10] (1,1);
    \draw[line width=0.8,line cap=round] (0,0.35) to [bend right=1] (0.23,0);
}}

\usepackage{newfloat}
\usepackage{listings}
\DeclareCaptionStyle{ruled}{labelfont=normalfont,labelsep=colon,strut=off} 
\floatstyle{ruled}
\newfloat{listing}{tb}{lst}{}
\floatname{listing}{Listing}
\title{Bi-level Contrastive Learning for Knowledge-Enhanced Molecule Representations}
\author {
    Pengcheng Jiang\textsuperscript{\rm 1},
    Cao Xiao\textsuperscript{\rm 2},
    Tianfan Fu\textsuperscript{\rm 3},
    Parminder Bhatia\textsuperscript{\rm 2},
    Taha Kass-Hout\textsuperscript{\rm 2},\\
    Jimeng Sun\textsuperscript{\rm 1},
    Jiawei Han\textsuperscript{\rm 1}
}
\affiliations {
    \textsuperscript{\rm 1}University of Illinois Urbana Champaign \quad
    \textsuperscript{\rm 2}GE HealthCare \quad
    \textsuperscript{\rm 3}Rensselaer Polytechnic Institute
}

\begin{document}

\maketitle

\begin{abstract}
Molecular representation learning is vital for various downstream applications, including the analysis and prediction of molecular properties and side effects. While Graph Neural Networks (GNNs) have been a popular framework for modeling molecular data, they often struggle to capture the full complexity of molecular representations. In this paper, we introduce a novel method called \textsc{Gode}, which accounts for the dual-level structure inherent in molecules. Molecules possess an intrinsic graph structure and simultaneously function as nodes within a broader molecular knowledge graph. \textsc{Gode} integrates individual molecular graph representations with multi-domain biochemical data from knowledge graphs. By pre-training two GNNs on different graph structures and employing contrastive learning, \textsc{Gode} effectively fuses molecular structures with their corresponding knowledge graph substructures. This fusion yields a more robust and informative representation, enhancing molecular property predictions by leveraging both chemical and biological information. When fine-tuned across 11 chemical property tasks, our model significantly outperforms existing benchmarks, achieving an average ROC-AUC improvement of 12.7\% for classification tasks and an average RMSE/MAE improvement of 34.4\% for regression tasks. Notably, \textsc{Gode} surpasses the current leading model in property prediction, with advancements of 2.2\% in classification and 7.2\% in regression tasks.
\end{abstract}

%

\section{Introduction}
In recent years, there has been a significant focus on tailoring machine learning models specifically for chemical and biological data~\citep{wang2021chemical, li2022graph, somnath2021multi, wang2023scientific}. A key challenge in this field is developing effective representations of molecular structures, which are critical for achieving accurate predictions in subsequent tasks~\citep{yang2019analyzing, haghighatlari2020learning}. To address this challenge, graph neural networks (GNNs) have become a widely adopted tool for facilitating representation learning~\citep{li2021dgl, hu2019strategies}. However, the conventional approach of using molecular graphs as input for GNNs may inadvertently constrain their potential for generating robust and comprehensive representations.

Molecular data, including chemical and biological datasets, exhibits a wide range of representational complexities~\citep{tong2017multi, argelaguet2020mofa+}. On an individual level, molecules can naturally be represented as graphs, with atoms as nodes and bonds as edges. For collections of molecules, their interrelationships can be captured through knowledge graphs (KGs), where each molecule is represented as a unique node. Notable examples of such KGs include UMLS~\citep{bodenreider2004unified}, PrimeKG~\citep{chandak2022building}, and PubChemRDF~\citep{fu2015pubchemrdf}. Building on this observation, we hypothesize that by integrating the molecular graphs of individual molecules and the broader sub-graphs from KGs centered on these molecules, we can create a more enriched representation that could lead to more accurate and robust predictions.

Previous efforts have attempted to unify molecular structures with KGs for property prediction. For example, \citet{ye2021unified} combined molecule embeddings with static KG embeddings~\citep{bordes2013translating}. However, these integrations often fall short of capturing the local molecular information within the KG, leading to only marginal improvements in prediction accuracy. In contrast, \citet{fang2023knowledge} demonstrated the advantages of enhancing molecular representations through contrastive learning, supported by a specialized chemical element KG. This approach has shown more significant performance gains, underscoring the value of integrating KGs with molecular data. Our work seeks to explore novel methods for embedding biochemical knowledge graphs into molecular prediction models.

In this study, we introduce “\textbf{G}raph as a N\textbf{ode}” (\textsc{Gode}), a new approach specifically designed to pre-train GNNs for molecule predictions. Our approach incorporates bi-level self-supervised tasks that target both molecular structures and their corresponding sub-graphs within the knowledge graph. By combining this strategy with contrastive learning, \textsc{Gode} produces more robust embeddings, leading to improved predictions of molecular properties.

Our major contributions can be summarized as follows:
\begin{itemize}[leftmargin=*]
    \item \textbf{A new paradigm for molecule knowledge integration}. Our \textsc{Gode} method introduces a new approach to integrating molecular structures with their corresponding KGs. This method not only produces richer and more accurate molecular representations in our specific application but also has the potential to be extended to other domains.
    
    \item \textbf{More robust molecular embeddings}.  Achieving robust molecular representations is crucial for accurate and consistent predictions across diverse datasets. Our approach integrates information from multiple domains for the same molecule, leveraging shared knowledge across modalities to create more comprehensive representations. By utilizing bi-level self-supervised pre-training combined with contrastive learning, we significantly enhance the robustness and reliability of the embeddings, resulting in more precise molecular property predictions and a solid foundation for various applications.

    \item \textbf{Introducing a new molecular knowledge graph}.  We have developed MolKG, a comprehensive knowledge graph specifically tailored to molecular data. MolKG encapsulates extensive molecular information, enabling more advanced and knowledge-driven molecular analyses.

\end{itemize}

To evaluate \textsc{Gode}'s performance, we conducted experiments across 11 chemical property prediction tasks. We benchmarked \textsc{Gode} against state-of-the-art methods, including GROVER~\citep{rong2020selfsupervised}, MolCLR~\citep{wang2021chemical}, and KANO~\citep{fang2023knowledge}. Our results demonstrate that \textsc{Gode} consistently outperforms these baselines, achieving improvements of 12.7\% in classification tasks and 34.4\% in regression tasks for molecular property prediction.

\section{Related Works}
\noindent\textbf{Graph-based Molecular Representation Learning. }
Over the years, various streams of molecular representation methods have been proposed, including traditional fingerprint-based  approaches~\citep{rogers2010extended}, SMILES string methods~\cite{xu2024smiles}, and modern GNN methods~\citep{jin2017predicting, jin2018learning, zheng2019predicting}. While Mol2Vec~\citep{jaeger2018mol2vec} adopts a molecule interpretation akin to Word2Vec for sentences~\citep{mikolov2013word2vec}, it overlooks substructure roles in chemistry. In contrast, GNN-based techniques can overcome this limitation by capturing more insightful details from aggregated sub-graphs. This advantage yields enhanced representations for chemical nodes, bonds, and entire molecules~\citep{rong2020selfsupervised, hu2019strategies, wang2021chemical}. Consequently, our study adopts GNN as the foundational framework for representing molecules.  

\noindent\textbf{Biomedical Knowledge Graphs. }
Various biomedical/biochemical KGs were developed to capture interconnections among diverse entities such as genes, proteins, diseases, and drugs~\citep{fu2015pubchemrdf, bodenreider2004unified}. Notably, PubChemRDF~\citep{fu2015pubchemrdf} spotlights biochemical domains, furnishing machine-readable chemical insights encompassing structures, properties, activities, and bioassays. Its subdivisions (e.g., \textit{Compound}, \textit{Cooccurrence}, \textit{Descriptor}, \textit{Pathway}) amass comprehensive chemical information.
PrimeKG~\citep{chandak2023building} is another KG that provides a multimodal view of precision medicine. Our study has a complementary focus and constructs a molecule-centric KG from those base KGs for supporting molecule property prediction tasks.

\noindent\textbf{Molecular Property Predictions.}
We focus on molecular property prediction, an essential downstream task for chemical representation learning frameworks. Three main aspects of the molecular property attract researchers: quantum mechanics properties~\citep{yang2019analyzing, liao2019lanczosnet, gilmer2017neural}, physicochemical properties~\citep{shang2018edge, wang2019molecule, becigneul2020optimal}, and toxicity~\citep{xu2017deep, yuan2020structpool}. Most of the recent works on molecular predictions are based on GNN~\citep{duvenaud2015convolutional,mansimov2019molecular}. However, the methods mentioned only focus on chemical structures and do not consider inter-relations among chemicals and knowledge graphs, which could improve property prediction. 

\noindent\textbf{Contrastive Learning in Molecular Representation.} The rise of cross-modality contrastive learning~\citep{radford2021learning} has increasingly influenced molecular representation approaches. Pioneering studies, such as \citep{pmlr-v162-stark22a}, have successfully employed contrastive learning to merge 3D and 2D molecular representations. This technique has been applied across various domains, including chemical reactions~\citep{lee2021retcl, seidl2022improving}, natural language processing~\citep{su2022molecular, seidl2023enhancing}, microscopy imaging~\citep{sanchez2022contrastive}, and chemical element knowledge~\citep{fang2023knowledge}. Distinctively, our work harnesses contrastive learning to enable knowledge transfer between KGs and molecular structures.

\begin{figure*}[!t]
    \centering
    \includegraphics[width=\linewidth]{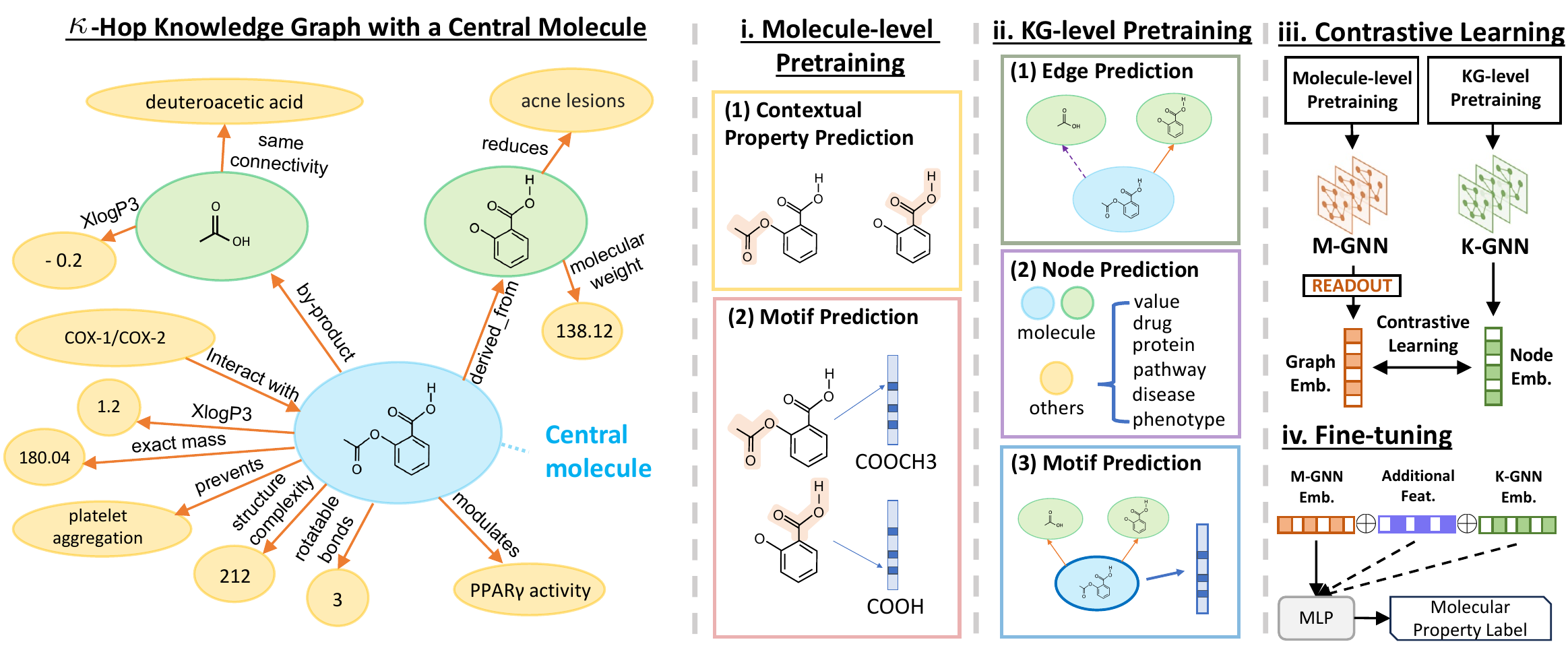}
    \caption{\textbf{Overview of our framework \textsc{Gode}}. \textbf{\textit{Left}}: The $\kappa$-hop KG sub-graph consisting of molecule-relevant relational knowledge, originating from a central molecule. \textbf{\textit{Right}}: We conduct (\textbf{i}) \textbf{Molecule-level Pre-training} on the molecular graphs with contextual property prediction and motif prediction tasks; (\textbf{ii}) \textbf{KG-level Pre-training} on the $\kappa$-hop KG sub-graphs of a central molecule with the tasks of edge prediction, node prediction, and motif prediction; (\textbf{iii}) \textbf{Contrastive Learning} to maximize the agreement between M-GNN and K-GNN, pre-trained by (i) and (ii), respectively; and (\textbf{iv}) \textbf{Fine-tuning} of our learned embedding, optionally enriched with extracted molecular-level features, for specific property predictions.
    }
    \label{fig:framework}
\end{figure*}
\noindent\textbf{Fusing Knowledge Graph and Molecules.} Previously, \citet{ye2021unified} introduced a method that combines static KG embeddings of drugs with their structural representations for downstream tasks. However, this approach overlooks the contextual information surrounding molecular nodes, leading to only modest performance improvements. Alternatively, \citet{wang2022imbalanced} proposed a Graph-of-Graphs technique that enriches molecular graph representations. While this method enhances the graph information, it does not explore pre-training or contrastive learning strategies to align the same entity across different graph modalities.
On the other hand, \citet{fang2023knowledge} pioneered a contrastive learning-based approach that augments molecular structures with element-wise knowledge, creating an innovative graph structure that has significantly improved molecular property predictions. Unlike existing methods, \textsc{Gode} extracts a molecule's sub-graph from our molecule-centric KG, offering a novel representation that effectively links molecular data with knowledge graphs.

\section{Method}
In this section, we present \textsc{Gode} framework. First, we define a few key concepts below.
\begin{definition}[\textbf{Molecule Graph}]
    A molecule graph (MG) is a structured representation of a molecule, where atoms (or nodes) are connected by bonds (or edges). An MG $G_m$ can be viewed as a graph structure with a set of nodes $\mathcal{V}_m$ representing atoms and a set of edges $\mathcal{E}_m$ representing bonds such that $G_m = (\mathcal{V}_m, \mathcal{E}_m)$.
\end{definition}

\begin{definition}[\textbf{Knowledge Graph}]
    A knowledge graph (KG) is a structured representation of knowledge in which entities (or nodes) are connected by relations (or edges). A directed KG can formally be represented as a set of $n$ triples: $\mathcal{T} = \{\leftb h, r, t \rightb_{i}\}_{i}^{n}$ where each triple contains a head entity ($h$) and a tail entity ($t$), and a relation ($r$) connecting them. A KG $G_k$ can also be viewed as a graph $G_k = (\mathcal{V}_k, \mathcal{E}_k)$ with a set of nodes $\mathcal{V}_k$ and a set of edges $\mathcal{E}_k$.
\end{definition}

\begin{definition}[\textbf{M-GNN}]
    M-GNN is a graph encoder $f: \mathcal{M} \rightarrow \mathbb{R}^{d}$ encoding a MG to a vector $\mathbf{h}_{\mathrm{MG}}$.
\end{definition}

\begin{definition}[\textbf{K-GNN}]
    K-GNN is a graph encoder $g: \mathcal{K} \rightarrow \mathbb{R}^d$ encoding the central molecule in a molecule KG sub-graph to a vector $\mathbf{h}_{\mathrm{KG}}$.
\end{definition}

\noindent  Our \textsc{Gode} approach (illustrated in Figure~\ref{fig:framework}) first conducts molecule-level pre-training to train an M-GNN and KG-level pre-training to train a K-GNN with a series of self-supervised tasks. Subsequently, we employ contrastive learning to enhance the alignment of molecule representations between the pre-trained M-GNN and K-GNN. Finally, we fine-tune our model for property prediction tasks.


\subsection{Molecule-level Pre-training}
\label{sub:molecule_level_pretrain}
Given a molecular graph $G_m = (\mathcal{V}_m, \mathcal{E}_m)$, 
we employ the GNN encoder to derive embeddings for atoms and bonds. To pre-train M-GNN, we employ two tasks described below.

\noindent \underline{(1) Node-level Contextual Property Prediction.}
We randomly select a node $v \in \mathcal{V}_m$ and its corresponding embedding $\mathbf{h}_v$. This embedding is then input into an output layer for predicting the contextual property. Contextual property prediction operates as a multi-class classification task. Here, the GNN's output layer computes the probability distribution for potential contextual property labels linked to node \(v\). These labels originate from the statistical attributes of the sub-graph centered on \(v\)~\citep{rong2020selfsupervised}.

\noindent \underline{(2) Graph-level Motif Prediction.}
The molecule graph embedding, represented as \(\mathbf{h}_{\mathrm{MG}}\), is also input into an output layer. This layer predicts the presence or absence of functional group motifs, which is detected by RDKit~\citep{landrum2013rdkit}. The embedding \(\mathbf{h}_{\mathrm{MG}}\) is derived by applying mean pooling to all nodes: $\mathbf{h}_{\mathrm{MG}} = {\mathrm{MEAN}}(\mathbf{h}_{v_1}, \mathbf{h}_{v_2}, ..., \mathbf{h}_{v_k} | v_1,v_2,...,v_k$ $\in \mathcal{V}_m)$, where \(\mathbf{h}_{v_1}\), \(\mathbf{h}_{v_2}\), ..., \(\mathbf{h}_{v_k}\) are the learned node embeddings from the M-GNN's final convolutional layer. This prediction task is a multi-label classification problem, where the GNN output layer predicts a binary label vector, indicating the presence or absence of each functional group motif in $G_m$.

During training, we employ a joint loss function, as shown in Eq.~\ref{eq:mgnn}, to optimize both the node-level contextual property prediction and the graph-level motif prediction. This loss function encourages the M-GNN to accurately predict both the contextual properties of nodes and the functional group motifs' presence or absence in MG.
\begin{align}
\label{eq:mgnn}
\mathcal{L}_{\mathrm{M}} = \sum_{v}^{\mathcal{V}'_m} \log P(p_v | \mathbf{h}_{v}) + \sum_{j=1}^{n} y_j \log P(M_j | \mathbf{h}_{\mathrm{MG}}) + \nonumber \\
(1-y_j) \log (1-P(M_j | \mathbf{h}_{\mathrm{MG}})), 
\end{align}
where $\mathcal{V}'_m$ is a set of randomly selected nodes; $p_v$ is the contextual property label for the node $v$; $n$ is the number of all possible motifs; $M_j$ is the presence of $j$-th motif.

After the molecule-level pre-training, M-GNN is able to encode a molecule to a vector $\mathbf{h}_{\mathrm{MG}}$ through mean pooling. 

\subsection{KG-level Pre-training}
\label{sub:kg_level_pretrain}
\underline{Embedding Initialization.} Prior to the K-GNN pre-training, we use knowledge graph embedding (KGE) methods~\citep{bordes2013translating, yang2014embedding, sun2019rotate, balavzevic2019tucker} to initialize the node and edge embeddings with entity and relation embeddings.  KGE methods capture relational knowledge behind the structure and semantics of entities and relationships in the KG. The KGE model is trained on the entire KG ($\mathcal{T}$) and learns to represent each entity and relation as continuous vectors in a low-dimensional space. The resulting embedding vectors capture the semantic meanings and relationships between entities and relations. The loss functions of KGE methods depend on the scoring functions they use. For example, TransE~\citep{bordes2013translating} learns embeddings for entities and relations in a KG by minimizing the difference between the sum of the head entity embedding ($\mathbf{e}_h$) and the relation embedding ($\mathbf{r}_r$), and the tail entity embedding ($\mathbf{e}_t$): $s(h, r, t) = -\lVert \mathbf{e}_h + \mathbf{r}_r - \mathbf{e}_t \rVert_{p}$, 
where $\lVert \cdot \rVert_p$ is the Lp norm.
After training the KGE model, we obtain the entity embeddings $\mathbf{e}_v$ and relation embeddings $\mathbf{r}_e$ for each node $v$ and edge $e$ in the KG, providing a strong starting point.

\noindent
\underline{Sub-graph Extraction.} for the central molecule is a crucial step in KG-level pre-training. Inspired by the work of G-Meta~\citep{huang2020graph}, we extract the sub-graph of each molecule to learn transferable knowledge from its surrounding nodes/edges in the biochemical KG. Specifically, for each central molecule, we extract a $\kappa$-hop sub-graph from the entire KG to capture its local neighborhood information. Given a molecule $m_i$, we first find its corresponding node $v_i$ in the KG, $G_k = (\mathcal{V}_k, \mathcal{E}_k)$. We then iteratively extract a neighborhood sub-graph $\mathcal{N}_k(v_i, h)$ of depth $h$ ($1 \leq h \leq \kappa$), centered at node $v_i$. The depth parameter $h$ determines the number of edge traversals to include in the sub-graph. To avoid over-smoothing, we terminate the expansion of a graph branch upon reaching a non-molecule node. Formally, the sub-graph extraction process is defined as follows. Let $\mathcal{N}_k(v, 0)$ be a single node $v$. For $h > 0$, $\mathcal{N}_k(v, h)$ is defined recursively as:
\begin{equation}
\label{eq:sub-graph}
\mathcal{N}_k(v, h) = \{v\} \cup \bigcup_{u \in \mathcal{N}_k(v, h-1)} \{u\} \cup \bigcup_{u \in \mathcal{M}} \{w : (u, w) \in \mathcal{E}_k\},
\end{equation}
where ${u}$ denotes the set of neighboring nodes of $v$ in the sub-graph $\mathcal{N}_k(v, h-1)$, and ${w : (u, w) \in \mathcal{E}_k}$ represents the set of nodes that share an edge with $u \in \mathcal{M}$ in the original KG $G_k$ where $\mathcal{M}$ is the set of molecule nodes. We define The $\kappa$-hop sub-graph for molecule $m$ is given by $G_{\textrm{sub}{(m, \kappa)}} = (\mathcal{V}_{\textrm{sub}{(m, \kappa)}}, \mathcal{E}_{\textrm{sub}{(m, \kappa)}}) = \mathcal{N}_k(c, \kappa)$ where $c$ is the corresponding node of $m$ in $G_{\textrm{sub}{(m, \kappa)}}$.
\\
The following loss function is used to pre-train K-GNN:
\begin{align}
\label{eq:kgnn_pretrain_tasks}
\mathcal{L}_{\mathrm{K}} = 
 &\lambda_{\mathrm{m}} \underbrace{\sum_{j=1}^{n} \mathrm{BCE}(y_j, P(M_j | \mathbf{h}_{c}))}_{\text{motif prediction}} + \lambda_{\text{n}} \underbrace{\mathrm{CE}(v', P(v' | \mathbf{h}_{v}))}_{\text{node prediction}} \nonumber \\
&+ \lambda_{\mathrm{e}} \underbrace{\mathrm{CE}((u,v)', P((u,v)' | \mathbf{h}_{u} \oplus \mathbf{h}_{v}))}_{\text{edge prediction}}
\end{align}
which includes three tasks shown in Figure~\ref{fig:framework} (ii): 
\begin{enumerate}[label={(\arabic*)}, leftmargin=*]
    \item \underline{Edge Prediction}, a multi-class classification task aiming at correctly predicting the edge type between two nodes: 
    \item \underline{Node Prediction}, a multi-class classification task predicting the category of a node in $G_{\textrm{sub}{(m, \kappa)}}$; 
    \item \underline{Node-level Motif Prediction}, a multi-label classification task predicting the motif of the central molecule node $c$ in $G_{\textrm{sub}{(m, \kappa)}}$. The motif labels are created by RDKit.
\end{enumerate}


\begin{table*}[!tp]
\small
\centering
\resizebox{\linewidth}{!}{
\begin{tabular}{lllll}

\toprule
\textbf{\# Triples}: 2,523,867 & \textbf{\# Entities}: 184,819 & \textbf{\# Relations}: 39 & \textbf{\# Entity Types}: 7 & \textbf{\# Molecules}: 65,454\\

\midrule
\multicolumn{5}{l}{\textbf{Entity Types}}\\
\multicolumn{5}{l}{\textit{molecule, \quad gene/protein, \quad disease, \quad effect/phenotype,\quad drug, \quad pathway, \quad value}}\\

\midrule
\multicolumn{5}{l}{\textbf{Relations}}\\

\multicolumn{5}{l}{\textit{drug\_protein,\quad contraindication,\quad indication,\quad off-label use,\quad drug\_drug,\quad drug\_effect,\quad defined\_bond\_stereo\_count,\quad tpsa, \quad rotatable\_bond\_count,}}\\

\multicolumn{5}{l}{\textit{xlogp3-aa,\quad structure\_complexity,\quad covalent\_unit\_count,\quad defined\_atom\_stereo\_count, \quad molecular\_weight,\quad hydrogen\_bond\_donor\_count,\quad}}\\

\multicolumn{5}{l}{\textit{undefined\_bond\_stereo\_count,\quad isotope\_atom\_count,\quad exact\_mass, \quad mono\_isotopic\_weight,\quad total\_formal\_charge,\quad hydrogen\_bond\_acceptor\_count,}}\\

\multicolumn{5}{l}{\textit{non-hydrogen\_atom\_count,\quad tautomer\_count, \quad undefined\_atom\_stereo\_count,\quad xlogp3,\quad cooccurence\_molecule\_molecule, \quad cooccurence\_molecule\_disease,}}\\

\multicolumn{5}{l}{\textit{cooccurence\_molecule\_gene/protein, \quad neighbor\_2d,\quad neighbor\_3d,\quad has\_same\_connectivity,\quad has\_component,\quad has\_isotopologue,\quad has\_parent,}}\\

\multicolumn{5}{l}{\textit{has\_stereoisomer,\quad to\_drug,\quad closematch,\quad type,\quad in\_pathway}}\\

\bottomrule
\end{tabular}
}
\caption{Overview of MolKG, a biochemical dataset we construct from PubChemRDF and PrimeKG. }
\label{tb:dataset}
\end{table*}

\noindent Here, ${(u,v)}'$ is the label of edge between the nodes $u$ and $v$. $v'$ is the label of node $v$, $\oplus$ denotes the embedding concatenation. $y_j$ is binary indicator, $P(M_j | \mathbf{h}_{c})$ is the predicted probability of central molecule $c$ has the $j$-th functional group motif $M_j$ given its embedding $\mathbf{h}_{c}$. $\lambda_{\mathrm{e}}$, $\lambda_{\mathrm{m}}$, and $\lambda_{\mathrm{n}}$ are balancing hyperparameters. 

After the KG-level pre-training, K-GNN can encode a molecule to a vector $\mathbf{h}_{\mathrm{KG}}$ given its surrounding nodes.

\subsection{Contrastive Learning}
\label{sub:contrastive_learning}
Inspired by the success of previous works~\citep{radford2021learning, seidl2023enhancing, sanchez2022contrastive} that apply contrastive learning to transfer knowledge across different modalities, we follow their steps using InfoNCE as the loss function to conduct contrastive learning between molecule graph and KG sub-graph. We construct the training set $\mathcal{D} = \mathcal{D}^{+} \cup \mathcal{D}^{-} = \{(m_i, s_i), y_i\}_N$, where $\mathcal{D}^{+} = \{(m_i, G_{\mathrm{sub}(m_i, \kappa)}), y_i = 1\}_{N_p}$ is a set of positive samples and $\mathcal{D}^{-} = \{(m_i, G_{\mathrm{sub}(m_j, \kappa)})_{j \neq i}, y_i = 0\}_{N-N_p}$ is a set of negative samples. To make the task more challenging, we further divide $\mathcal{D}^{-}$ into $\mathcal{D}^{-}_{rand}$, and $\mathcal{D}^{-}_{nbr}$, which are (1) randomly sampled from all negative molecule-centric KG sub-graphs , and (2) sampled from the sub-graphs of the neighbor molecule nodes connected to the positive molecule node, respectively. The loss is defined as:
\begin{align}
\label{eq:infonce}
    \mathcal{L}_{\mathrm{InfoNCE}} = -\frac{1}{N} \sum_{i=1}^{N} \biggl[ y_i \log(\mathrm{sim}(f(m_i), g(s_i))) \nonumber\\
    + (1-y_i)\log(1-\mathrm{sim}(f(m_i), g(s_i)))\biggr],
\end{align}
where $\mathrm{sim}(f(m_i), g(s_i))) = \frac{\exp{(\tau^{-1}\mathbf{h}_{\mathrm{MG}(i)}^{\mathrm{T}}\mathbf{h}_{\mathrm{KG}(i)})}}{\exp{(\tau^{-1}\mathbf{h}_{\mathrm{MG}(i)}^{\mathrm{T}}\mathbf{h}_{\mathrm{KG}(i)})} + 1}$, $y_i$ is the binary label, $m_i$ and $s_i$ are the paired MG and KG sub-graph in the training data, $\tau^{-1}$ is the inverse temperature.

\subsection{Fine-tuning for Downstream Tasks}
\label{sub:finetuning}
Upon completing molecule- and KG-level pre-training combined with contrastive learning, we obtain two GNN encoders, $f$ and $g$, which respectively encode molecules and KG sub-graphs into vectors. 
Following previous works \citep{rong2020selfsupervised, fang2023knowledge, wu2018moleculenet, yang2019analyzing}, we employ RDKit to extract additional molecule-level features $\mathbf{h}_\mathrm{f}$. A joint representation is formed by $\mathbf{h}_{\mathrm{joint}} = \mathbf{h}_{\mathrm{MG}} \oplus \mathbf{h}_\mathrm{f} \oplus \mathbf{h}_{\mathrm{KG}}$, with $\oplus$ representing concatenation. This representation is then utilized to predict the target property $y$ using a multi-layer perception (MLP) with an activation function. For multi-label classification, we use Binary Cross-Entropy (BCE) loss with sigmoid activation, and for regression, we use Mean Squared Error (MSE) loss.

\section{Experiments}
\subsection{Experimental Setting}
\textbf{Data Sources.} (1) \underline{Molecule-level pre-training data}: The pre-training data for our molecule-level M-GNN is derived from the same unlabelled dataset of 11 million molecules utilized by GROVER. This dataset encompasses sources such as ZINC15~\citep{sterling2015zinc} and ChEMBL~\citep{gaulton2012chembl}. We randomly split this dataset into two subsets with a 9:1 ratio for training and validation.
(2) \underline{Knowledge graph-level pre-training data}: For the KG-level pre-training, we retrieve KG triples related to the molecules from PubChemRDF and PrimeKG. These include various subdomains and properties from PubChemRDF, as well as 3-hop sub-graphs for all 7957 drugs from PrimeKG. We show an overview of the dataset (MolKG) in Table~\ref{tb:dataset}. The dataset is divided into training and validation sets with a 9:1 ratio. \textit{The construction details of the dataset are placed in Appendix}. (3) \underline{Contrastive learning data}: we set the negative sampling ratio as  $ |\mathcal{D}^{-}|/|\mathcal{D}^{+}|=32$ and retain a $1:1$ ratio for $\mathcal{D}^{-}_{rand}: \mathcal{D}^{-}_{nbr}$. Training and validation samples are in a $0.95:0.05$ ratio.
(4) \underline{Downstream task datasets}: The effectiveness of our model is tested utilizing the comprehensive MoleculeNet dataset~\citep{wu2018moleculenet,huang2021therapeutics}\footnote{\url{https://moleculenet.org/datasets-1}}, which contains 6 classification and 5 regression datasets for molecular property prediction. We place detailed descriptions of these datasets in the Appendix.
To fine-tune the model, we calculate the mean and standard deviation of the ROC-AUC for classification tasks and RMSE/MAE for regression tasks. Scaffold
splitting with three random seeds was employed with a training/validation/testing ratio of 8:1:1 across all datasets, aligning with previous studies~\citep{rong2020selfsupervised, fang2023knowledge}.

\noindent\textbf{Baselines.} We compare our proposed model with several popular baselines in molecular property prediction tasks, which include GCN~\citep{kipf2016semi}, GIN~\citep{xu2018powerful}, SchNet~\citep{schutt2017schnet}, MPNN~\citep{gilmer2017neural}, DMPNN~\citep{yang2019analyzing}, MGCN~\citep{lu2019molecular}, N-GRAM~\citep{liu2019n}, Hu et al~\citep{hu2019strategies}, GROVER~\citep{rong2020selfsupervised}, MGSSL~\citep{zhang2021motifbased}, KGE\_NFM~\citep{ye2021unified} with our MolKG, MolCLR~\citep{wang2102molclr}, and KANO~\citep{fang2023knowledge}.

\noindent\textbf{Implementation.}
For molecule-level pre-training, we employ GROVER~\citep{rong2020selfsupervised}, and for KG-level pre-training, we utilize GINE~\citep{hu2019strategies}. TransE initializes the KG embeddings over a span of 10 epochs. Our settings include $\lambda_{\textrm{e}}=1.5$, $\lambda_{\textrm{m}}=1.8$, and $\lambda_{\textrm{n}}=1.5$. Both M-GNN and K-GNN have a hidden size of 1,200. We adopt a temperature $\tau = 1.0$ for contrastive learning. Early stopping is anchored to validation loss. During fine-tuning, embeddings from K-GNN remain fixed, updating only the parameters of M-GNN.
We use Adam optimizer with the Noam learning rate scheduler~\citep{vaswani2017attention}. All tests are performed with two AMD EPYC 7513 32-core Processors, 528GB RAM, 8 NVIDIA A6000 GPUs, and CUDA 11.7. \textit{More implementation details and the hyper-parameter study are placed in Appendix}.

\begin{table*}[t]
\centering
\resizebox{\textwidth}{!}{
\begin{tabular}{l|rrrrrr|rrrrr}
\toprule
& \multicolumn{6}{c|}{\textbf{Classification (Higher is Better)}} & \multicolumn{5}{c}{\textbf{Regression (Lower is Better)}} \\ 
\midrule
\textbf{Dataset} & \textbf{BBBP} & \textbf{SIDER} & \textbf{ClinTox} & \textbf{BACE} & \textbf{Tox21} & \textbf{ToxCast} & \textbf{FreeSolv} & \textbf{ESOL} & \textbf{Lipophilicity} & \textbf{QM7} & \textbf{QM8} \\
\# Molecules   & 2039  & 1427  &1478    &1513  &7831   &8575 &642 &1128 &4200 &6830 &21786\\
\# Tasks       &1      &27     &2       &1     &12     &617 & 1 &1 &1 &1 &12\\
\midrule\midrule
GCN     
& $71.8 \pm 0.9$ 
& $53.6 \pm 0.3$ 
& $62.5 \pm 2.8$ 
& $71.6 \pm 2.0$ 
& $70.9 \pm 0.3$
& $65.0 \pm 6.1$ 
& $2.870 \pm 0.140$
& $1.430 \pm 0.050$
& $0.712 \pm 0.049$
& $122.9 \pm 2.2$
& $0.037 \pm 0.001$
\\
GIN               
& $65.8 \pm 4.5$ 
& $57.3 \pm 1.6$ 
& $58.0 \pm 4.4$ 
& $70.1 \pm 5.4$ 
& $74.0 \pm 0.8$ 
& $66.7 \pm 1.5$ 
& $2.765 \pm 0.180$
& $1.452 \pm 0.020$
& $0.850 \pm 0.071$
& $124.8 \pm 0.7$
& $0.037 \pm 0.001$
\\
SchNet           
& $84.8 \pm 2.2$ 
& $54.5 \pm 3.8$ 
& $71.7 \pm 4.2$ 
& $76.6 \pm 1.1$
& $76.6 \pm 2.5$ 
& $67.9 \pm 2.1$  
& $3.215 \pm 0.755$
& $1.045 \pm 0.064$
& $0.909 \pm 0.098$
& $74.2 \pm 6.0$
& $0.020 \pm 0.002$
\\
MPNN         
& $91.3 \pm 4.1$
& $59.5 \pm 3.0$
& $87.9 \pm 5.4$ 
& $81.5 \pm 4.4$ 
& $80.8 \pm 2.4$
& $69.1 \pm 1.3$
& $1.621 \pm 0.952$
& $1.167 \pm 0.430$
& $\textbf{0.672} \pm \textbf{0.051}$
& $111.4 \pm 0.9$
& $\textbf{0.015} \pm \textbf{0.001}$
\\
DMPNN         
& $91.9 \pm 3.0$ 
& $63.2 \pm 2.3$
& $89.7 \pm 4.0$ 
& $85.2 \pm 5.3$
& $\textbf{82.6} \pm \textbf{2.3}$
& $71.8 \pm 1.1$ 
& $1.673 \pm 0.082$
& $1.050 \pm 0.008$
& $0.683 \pm 0.016$
& $103.5 \pm 8.6$
& $\textbf{0.016} \pm \textbf{0.001}$
\\
MGCN            
& $85.0 \pm 6.4$ 
& $55.2 \pm 1.8$ 
& $63.4 \pm 4.2$ 
& $73.4 \pm 3.0$
& $70.7 \pm 1.6$ 
& $66.3 \pm 0.9$
& $3.349 \pm 0.097$
& $1.266 \pm 0.147$
& $1.113 \pm 0.041$
& $77.6 \pm 4.7$
& $0.022 \pm 0.002$
\\
N-GRAM                 
& $91.2 \pm 1.3$ 
& $63.2 \pm 0.5$
& $85.5 \pm 3.7$ 
& $87.6 \pm 3.5$
& $76.9 \pm 2.7$ 
& -  
& $2.512 \pm 0.190$
& $1.100 \pm 0.160$
& $0.876 \pm 0.033$
& $125.6 \pm 1.5$
& $0.032 \pm0.003$
\\
HU. et.al        
& $70.8 \pm 1.5$
& $62.7 \pm 0.8$
& $72.6 \pm 1.5$ 
& $84.5 \pm 0.7$
& $78.7 \pm 0.4$ 
& $65.7 \pm 0.6$ 
& $2.764 \pm 0.002$
& $1.100 \pm 0.006$
& $0.739 \pm 0.003$
& $113.2 \pm 0.6$
& $0.022 \pm 0.001$
\\
\rowcolor{gray!20} $\textrm{GROVER}_{\textrm{Large, GTrans}}$     
& $86.2 \pm 3.9$
& $57.6 \pm 1.6$
& $74.7 \pm 4.4$
& $82.5 \pm 4.4$
& $76.9 \pm 2.3$
& $66.7 \pm 2.6$
& $2.445 \pm 0.761$
& $1.028 \pm 0.145$
& $0.890 \pm 0.050$
& $95.3 \pm 5.6$
& $0.020 \pm 0.003$
\\
MGSSL           
& $70.5 \pm 1.1$
& $64.1 \pm 0.7$
& $80.7 \pm 2.1$
& $79.7 \pm 0.8$
& $76.4 \pm 0.4$
& $64.1 \pm 0.7$
& - & - & - & - & -
\\
MolCLR
& $73.3 \pm 1.0$
& $61.2 \pm 3.6$
& $89.8 \pm 2.7$
& $82.8 \pm 0.7$
& $74.1 \pm 5.3$
& $65.9 \pm 2.1$
& $2.301 \pm 0.247$
& $1.113 \pm 0.023$
& $0.789 \pm 0.009$
& $90.0 \pm 1.7$
& $0.019 \pm 0.013$
\\
\rowcolor{yellow!10} $\text{MolCLR}_{\text{GTrans}}$
& $76.7 \pm 2.2$
& $63.3 \pm 2.5$
& $89.3 \pm 3.1$
& $87.7 \pm 1.8$
& $80.2 \pm 3.2$
& $70.4 \pm 2.1$
& $2.124 \pm 0.223$
& $0.982 \pm 0.109$
& $0.767 \pm 0.064$
& $88.9 \pm 4.8$
& $0.018 \pm 0.002$
\\
\midrule
\rowcolor{yellow!10} $\text{KGE\_NFM}_{\text{w/ MolKG}}$
& $92.4 \pm 2.4$
& $\textbf{65.3} \pm \textbf{1.4}$
& $87.3 \pm 2.0$ 
& $78.1 \pm 2.1$
& $79.8 \pm 3.3$ 
& $\textbf{72.6} \pm \textbf{1.8}$ 
& $1.942 \pm 0.441$
& $1.027 \pm 0.201$
& $0.877 \pm 0.071$
& $87.6 \pm 3.2$
& $\textbf{0.016} \pm \textbf{0.001}$
\\
$\text{KANO}_{\text{CMPNN}}$
& $\textbf{92.6} \pm \textbf{1.8}$
& $\textbf{65.5} \pm \textbf{1.6}$
& $\textbf{92.9} \pm \textbf{1.1}$
& $\textbf{90.7} \pm \textbf{3.1}$
& $81.8 \pm 1.1$
& $\textbf{72.5} \pm \textbf{1.9}$
& $\textbf{1.320} \pm \textbf{0.244}$
& $\textbf{0.902} \pm \textbf{0.104}$
& ${\textbf{0.641} \pm \textbf{0.012}}$
& $\textbf{66.5} \pm \textbf{3.7}$
& ${\textbf{0.013} \pm \textbf{0.001}}$
\\
\rowcolor{yellow!10} $\text{KANO}_{\text{GTrans}}$
& $\textbf{93.7} \pm \textbf{2.3}$
& $63.8 \pm 1.2$
& $\textbf{93.6} \pm \textbf{0.7}$
& $\textbf{90.4} \pm \textbf{1.5}$
& $\textbf{81.2} \pm \textbf{1.8}$
& $\textbf{72.5} \pm \textbf{1.5}$
& $\textbf{1.443} \pm \textbf{0.315}$
& $\textbf{0.914} \pm \textbf{0.092}$
& $\textbf{0.651} \pm \textbf{0.018}$
& $\textbf{63.6} \pm \textbf{4.1}$
& ${\textbf{0.013} \pm \textbf{0.002}}$
\\ \midrule
\rowcolor{yellow!10} \textbf{\textsc{Gode} (ours)}
& ${\textbf{94.8} \pm \textbf{1.9}}$
& ${\textbf{67.4} \pm \textbf{1.4}}$
& ${\textbf{94.7} \pm \textbf{2.9}}$
& ${\textbf{92.0} \pm \textbf{2.2}}$
& ${\textbf{84.3} \pm \textbf{1.2}}$
& ${\textbf{73.4} \pm \textbf{0.9}}$
& ${\textbf{1.048} \pm \textbf{0.314}}$
& ${\textbf{0.746} \pm \textbf{0.128}}$
& $0.743 \pm 0.043$
& ${\textbf{57.2} \pm \textbf{3.0}}$
& ${\textbf{0.013} \pm \textbf{0.001}}$
\\
\bottomrule
\end{tabular}
}
\caption{Performance on six classification benchmarks (ROC-AUC; higher is better) and five regression benchmarks (RMSE for FreeSolv, ESOL, and Lipophilicity; MAE for QM7/QM8; lower is better). We report the mean and standard deviation. The Top-3 results are highlighted in bold. The backbone model is shaded in grey, and models utilizing the backbone are shaded in yellow. The table is divided into three sections: non-KG methods, other KG-based methods, and our method.}
\label{tab:combined_results}
\end{table*}
\begin{figure*}[htbp]
    \centering
    \includegraphics[width=\linewidth]{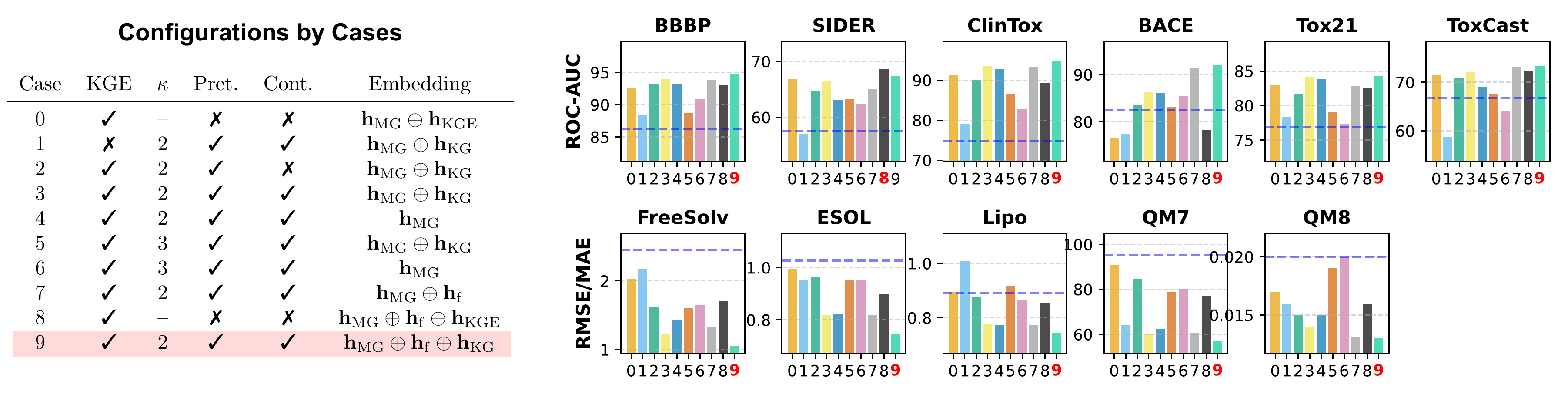}
    \caption{Ablation study configurations and results. (Left) Configurations. ``KGE'': KG embedding initialization. ``$\kappa$'': $\kappa$-hop KG subgraph. ``Pret.'': KG-level pre-training. ``Cont.'': contrastive learning. ``Embedding'': input to MLP for fine-tuning. (Right) Performance comparison across different datasets and configurations. We highlight the best configuration for each dataset in red. The dotted blue lines denote the performance achieved by the backbone model (GROVER).}
  \label{fig:ablation_study}
\end{figure*}

\subsection{Results on Molecule Property Prediction}
Table \ref{tab:combined_results}
presents comparative performance metrics for classification and regression tasks, respectively. It is clear from the data that our proposed method, \textsc{Gode}, consistently outperforms the baseline models in most tasks. Specifically, in classification tasks, \textsc{Gode} achieves SOTA results across all tasks. Amongst the competitors, KANO stands out, consistently showcasing performance close to our method. Intriguingly, KANO, as a knowledge-driven model, augments molecular structures by integrating information about chemical elements from its ElementKG. This underlines the substantial advantage of leveraging external knowledge in predicting molecular properties. On the regression front, \textsc{Gode} attains best results in 4 out of 5 tasks. This consistent high performance, irrespective of the nature of the task, underscores our model's adaptability and reliability. Cumulatively, our approach yields a relative improvement of 23.6\% across all tasks (12.7\% for classification and 34.4\% for regression tasks). When compared with the SOTA model, KANO, \textsc{Gode} records improvements of 2.2\% and 7.2\% for classification and regression tasks, respectively. 

To analyze the effects of \textsc{Gode}'s variants, we conduct ablation studies in Figure~\ref{fig:ablation_study}, which are discussed as follows.

\noindent\textbf{Effect of the Integration of MolKG.} To assess the impact of integrating our molecule-centric KG - MolKG, into molecule property prediction, we compare Case {8} in Figure~\ref{fig:ablation_study} with our backbone M-GNN model, GROVER. Specifically, Case {8} melds GROVER ($\mathbf{h}_\mathrm{MG} \oplus \mathbf{h}_\mathrm{f}$) with the static KG embedding ($\mathbf{h}_\mathrm{KGE}$), which is trained using the KGE method. Our observations indicate that infusing the KG boosts performance across all tasks, resulting in a noteworthy 14.3\% overall enhancement. Moreover, when all variants of \textsc{Gode} are deployed (as in Case {9}), a significant uplift of 23.2\% in performance over GROVER is realized. 


\begin{table*}[t]
\centering
\resizebox{\textwidth}{!}{
\begin{tabular}{cc|cccccc|ccccc}
\toprule
\multicolumn{13}{c}{\textbf{(a) Effect of Pre-training on M-GNN and K-GNN}} \\
\midrule
M-GNN Pret. & K-GNN Pret. & BBBP & SIDER & ClinTox & BACE & Tox21 & ToxCast & FreeSolv & ESOL & Lipo & QM7 & QM8 \\
\midrule
\cmark & \cmark & \textbf{94.8} & \textbf{67.4} & \textbf{94.7} & \textbf{92.0} & \textbf{84.3} & \textbf{73.4} & \textbf{1.048} & \textbf{0.746} & 0.743 & \textbf{57.2} & \textbf{0.013} \\
\xmark & \cmark & 92.2 & 62.6 & 89.4 & 89.8 & 80.6 & 70.8 & 1.313 & 0.834 & \textbf{0.708} & 64.6 & 0.016 \\
\cmark & \xmark & 93.2 & 66.7 & 90.7 & 81.6 & 83.1 & 71.9 & 1.563 & 0.841 & 0.876 & 74.4 & 0.017 \\
\xmark & \xmark & 88.9 & 62.1 & 88.4 & 84.1 & 81.6 & 69.4 & 1.944 & 0.978 & 0.845 & 77.9 & 0.017 \\
\midrule
\multicolumn{13}{c}{\textbf{(b) Effect of Relationship Exclusion from MolKG}} \\
\midrule
\multicolumn{2}{c|}{Knowledge Graph} & BBBP & SIDER & ClinTox & BACE & Tox21 & ToxCast & FreeSolv & ESOL & Lipo & QM7 & QM8 \\
\midrule
\multicolumn{2}{l|}{MolKG} & 94.8 & \textbf{67.4} & \textbf{94.7} & \textbf{92.0} & \textbf{84.3} & \textbf{73.4} & \textbf{1.048} & \textbf{0.746} & \textbf{0.743} & \textbf{57.2} & 0.013 \\
\multicolumn{2}{l|}{w/o \textit{indication}} & 93.8 & 65.7 & 93.4 & 91.6 & 84.2 & 73.0 & 1.063 & 0.754 & 0.751 & 58.1 & 0.013 \\
\multicolumn{2}{l|}{w/o \textit{xlogp3} \& \textit{xlogp3-aa}} & 93.7 & 66.0 & 94.2 & 91.1 & 83.0 & 72.8 & 1.189 & 0.789 & 0.782 & 57.8 & \textbf{0.012} \\
\multicolumn{2}{l|}{w/o \textit{tautomer\_cnt} \& \textit{covalent\_unit\_cnt}} & 94.3 & 66.5 & 93.1 & 90.9 & 83.5 & 72.5 & 1.272 & 0.761 & 0.759 & 61.7 & 0.014 \\ 
\multicolumn{2}{l|}{w/o \textit{nbr\_2d} \& \textit{nbr\_3d} \& \textit{has\_same\_conn}} & \textbf{95.0} & 67.3 & 93.6 & 91.3 & \textbf{84.3} & 72.7 & 1.058 & 0.749 & 0.748 & 57.6 & 0.013 \\
\bottomrule
\end{tabular}
}
\caption{Study the effects of (top) bi-level self-supervised pre-training and (below) relationship exclusion from MolKG.} 
\label{tb:combined_performance}
\end{table*}
\noindent\textbf{Effect of KG-level Pre-training and Contrastive Learning.}
Through a side-by-side comparison of Cases {0}, {2}, and {3} in Figure~\ref{fig:ablation_study}, we discern the value of K-GNN pre-training and contrastive learning. 
Standalone K-GNN pre-training (Case {2}) yields a modest boost of 4.5\%, with a particularly slight edge in classification tasks at 0.1\%. However, when paired with contrastive learning and leveraging both $\mathbf{h}_\mathrm{MG}$ and $\mathbf{h}_\mathrm{KG}$ for fine-tuning, as in Case {3}, the surge in performance is notable, reaching an overall enhancement of 13.6\% over the baseline Case {0}.
A testament to the effectiveness of this approach can be seen in the BBBP dataset. The molecule acetylsalicylate, better known as aspirin, posed a prediction challenge to both our M-GNN model and the methods in Cases {0} and {2}. Yet, when Case {3} employed relational knowledge from its KG sub-graph (e.g., [\textit{acetylsalicylate, indication, neurological conditions}]) alongside contrastive learning, it managed to make accurate predictions. This example underscores the pivotal role of contrastive learning in refining molecular property predictions.


\noindent\textbf{Efficacy of Knowledge Transfer.}
The influence of contrastive learning in transferring domain knowledge from the biochemical KG to the molecular representation $\mathbf{h}_\mathrm{MG}$ is discerned by examining Cases 3, 4, 5, 6, and contrasting GROVER (backbone) with Cases {7} and {9}. Notably, while the M-GNN embeddings of \textsc{Gode} (represented by Cases {4} and {6}) do not quite surpass the bi-level concatenated embeddings (Cases {3} and {5}), they come notably close. More compelling is Case {7}, which parallels Case {9} and outperforms GROVER by a striking 21.0\% (with 12.0\% in classification and 30.1\% in regression). 
The distinguishing feature of Case {7} that provides an edge over GROVER is its enriched $\mathbf{h}_\mathrm{MG}$, an enhancement absent in GROVER. This underscores \textsc{Gode}'s prowess in biochemical knowledge transfer to molecular representations.

\noindent\noindent\textbf{Mutual Benefit of Bi-level Self-supervised Pre-training.}
In addition to the insights provided in Figure \ref{fig:ablation_study}, we conducted an in-depth analysis of the impact of pre-training M-GNN and K-GNN on the performance of \textsc{Gode}, as detailed in Table \ref{tb:combined_performance}(a). The findings clearly underscore the mutual benefit of both M-GNN and K-GNN pre-training to the efficacy of our framework. Notably, an improved performance on the Lipophilicity dataset was observed when the M-GNN pre-training was omitted, presenting an intriguing aspect for further investigation.

\noindent\textbf{Impact of Relationship Exclusion.} We investigated the impact of removing specific relationships from MolKG on both classification and regression datasets (in Table \ref{tb:combined_performance}(b)). For classification, excluding ``tautomer\_count'' and ``covalent\_unit\_count'' led to the largest performance drop on ClinTox, while removing structural similarity relationships slightly improved results on BBBP. For regression, removing ``xlogp3'' and ``xlogp3-aa'' substantially increased the error on solvation and lipophilicity predictions, aligning with the physical meaning of these features. Removing ``tautomer\_count'' and ``covalent\_unit\_count'' also notably impacted FreeSolv and QM7, suggesting their importance for predicting solvation and quantum properties. 
This analysis reveals the variable significance of different relationships, with the most consistent impact observed for ``xlogp3'' and ``xlogp3-aa'' on solvation and lipophilicity tasks.

\begin{figure}[!t]
  \centering
  \includegraphics[width=1.0\linewidth]{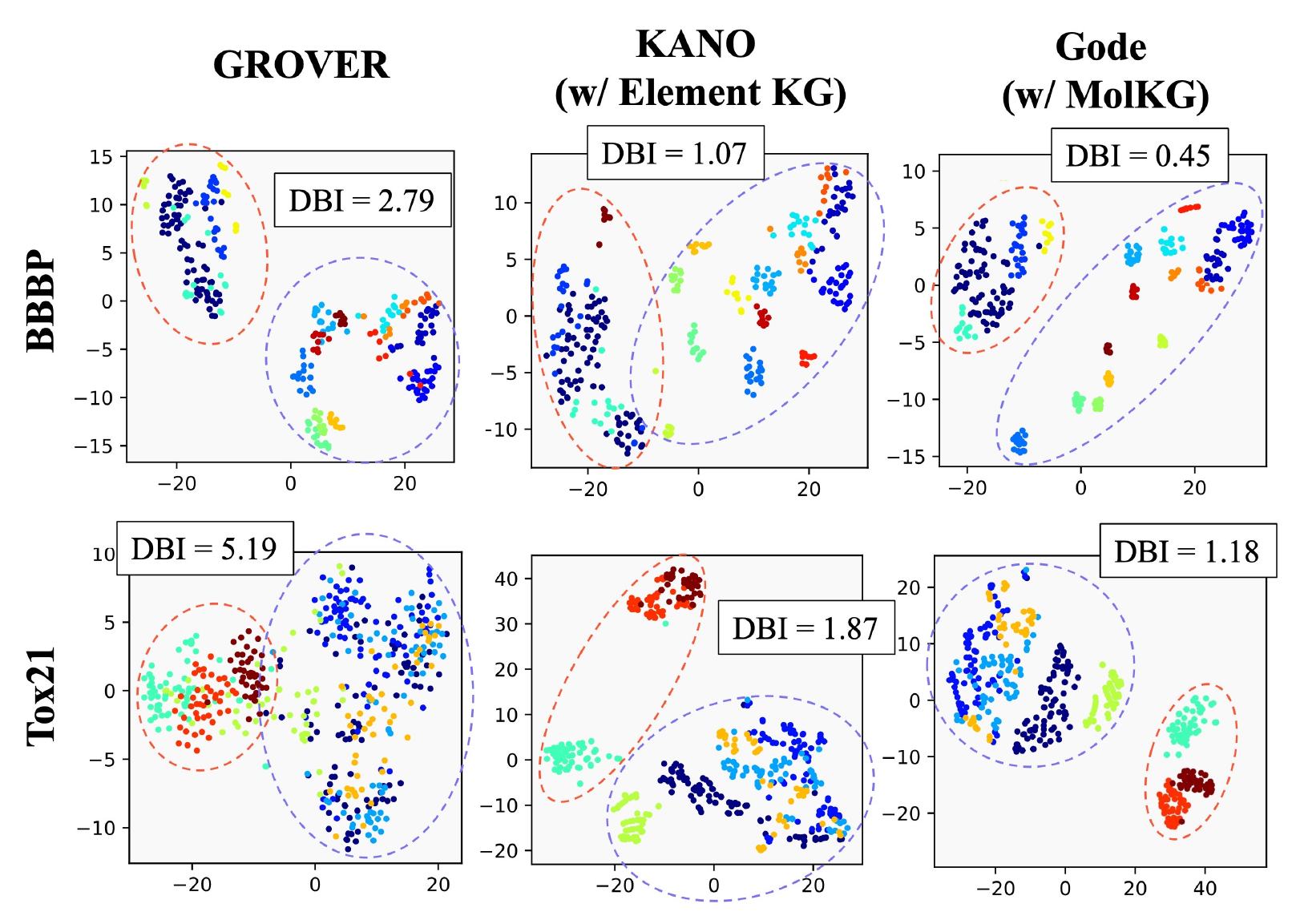}
  \caption{t-SNE visualization of molecule embeddings across two tasks. Each color represents a unique scaffold (molecule substructure). We compare the embeddings from GROVER, KANO, and \textsc{Gode}. The clustering quality is assessed using the DB index.}
  \label{fig:tsne}
\end{figure}

\noindent\textbf{Embedding Visualization.} In the t-SNE visualization presented in Figure \ref{fig:tsne}, the GROVER embeddings highlight molecules from varying scaffolds intermingling, signaling a significant avenue for refinement. Particularly in the Tox21 task, these embeddings appear sparse. When enhanced with KANO, there is a noticeable delineation of clusters, reflecting the constructive influence of integrating external knowledge into molecular representations. Nonetheless, a residual overlap of molecules from different scaffolds still persists. Progressing to the \textsc{Gode} visualization, the clusters exhibit further refinement, achieving pronounced distinctiveness with minimal scaffold overlap, outperforming KANO, and securing the lowest Davies–Bouldin Index (DBI), which signifies the effectiveness of \textsc{Gode}.


\section{Conclusion}
We introduced \textsc{Gode}, a framework that enhances molecule representations through bi-level self-supervised pre-training and contrastive learning, leveraging biochemical domain knowledge. Our empirical results demonstrate its effectiveness in molecular property prediction tasks. Future work will focus on expanding the coverage of MolKG and identifying crucial knowledge elements for optimizing molecular representations. This research lays groundwork for advancements in drug discovery applications.

\section{Acknowledgments}
Research was supported in part by US DARPA INCAS Program No. HR0011-21-C0165 and BRIES Program No. HR0011-24-3-0325, National Science Foundation IIS-19-56151, the Molecule Maker Lab Institute: An AI Research Institutes program supported by NSF under Award No. 2019897, and the Institute for Geospatial Understanding through an Integrative Discovery Environment (I-GUIDE) by NSF under Award No. 2118329. 

\bibliography{references}
\clearpage
\appendix
\section{Broader Impact}
The development of \textsc{Gode} offers a significant advance in the realm of molecular representation learning. Its broader impacts can be summarized as follows:

\noindent \textbf{Enhanced Drug Discovery} By providing a robust representation of molecules enhanced by knowledge, \textsc{Gode} can potentially accelerate drug discovery processes. This could lead to faster identification of potential drug candidates and reduce the time and cost associated with the introduction of new drugs into the market.

\noindent \textbf{Interdisciplinary Applications} The fusion of molecular structures with knowledge graphs can be applied beyond the realm of molecular biology. This approach can be extended to other scientific domains where entities have both intrinsic structures and are part of larger networks.

\noindent \textbf{Potential Ethical Considerations} As with any predictive model, there is a need to ensure that the data used is unbiased and representative. Misrepresentations or biases in the knowledge graph or molecular data can lead to skewed predictions, which could have implications in real-world applications, especially in drug development.

\section{MolKG Construction and Processing}
The construction of our molecule-centric knowledge graph - MolKG, involved a comprehensive data retrieval process of knowledge graph triples relevant to molecules. We retrieve the data from two distinguished sources: PubChemRDF\footnote{\url{https://pubchem.ncbi.nlm.nih.gov/docs/rdf-intro}}~\citep{fu2015pubchemrdf} and PrimeKG~\citep{chandak2022building}.
From PubChemRDF, we concentrated on triples from six specific subdomains:
\begin{itemize}[leftmargin=*]
    \item \textit{Compound}: This encompasses compound-specific relation types such as \textit{parent compound}, \textit{component compound}, and \textit{compound identity group}.
    \item \textit{Cooccurrence}: This domain captures triples like \textit{compound-compound}, \textit{compound-disease}, and \textit{compound} \textit{-gene} co-occurrences. By ranking co-occurrences based on their scores, we selected the top 5 compounds, diseases, and genes for each molecule, resulting in at most 15 co-occurred entities per molecule.
    \item \textit{Descriptor}: This domain details explicit molecular properties including \textit{structure complexity}, \textit{rotatable bond}, and \textit{covalent unit count}.
    \item \textit{Neighbors}: Represents the top $N$ molecules similar in 2D and 3D structures. For our dataset, we integrated the top 3 similar molecules from both 2D and 3D structures for each molecule.
    \item \textit{Component}: Associates molecules with their constituent components.
    \item \textit{Same Connectivity}: Showcases molecules with identical connectivity to source molecules.
\end{itemize}
From PrimeKG, we pursued a rigorous extraction technique, deriving 3-hop sub-graphs for all 7,957 drugs, regarded as molecules, from the entirety of the knowledge graph. Consistency and accuracy in data handling were paramount. We utilized recognized information retrieval tools\footnote{\url{https://pubchem.ncbi.nlm.nih.gov/docs/pug-rest}}\footnote{\url{https://www.ncbi.nlm.nih.gov/home/develop/api/}} to bridge various representations and coding paradigms for identical molecular entities. Compound ID (CID) served as our go-to medium for molecular conversions across the two knowledge graphs. 

Lastly, within our assembled knowledge graph, entities identified as ``value'' are normalized to (1, 10). Subsequently, we classified these entities, ensuring a maximum class count of 10. 

We attached the entire MolKG dataset\footnote{in \textit{gode\_data/data\_process/KG\_processed.csv}} and the detailed processing scripts for its construction\footnote{in \textit{gode\_data/dataset\_construction/}} as supplemental material.

\begin{table*}[!t]
\centering
\caption{Summary of hyper-parameter study for the experimental setup. We \textbf{highlight} the best setting used in experiments.}
\resizebox{0.8\linewidth}{!}{\begin{tabular}{ll}
\toprule
\textbf{Hyper-parameter} & \textbf{Studied Values} \\
\midrule
\textbf{M-GNN} \\
GNN model           & $\mathrm{GROVER}_{\mathrm{w/} \{\textbf{GTransformer}, \mathrm{MPNN}, \mathrm{GIN}\}}$ \\
learning rate       & 1.5e-4 \\
weight decay        & 1e-7 \\
hidden dimension    & \{400, 800, \textbf{1200}\} \\
pre-training epochs & 500 \\
dropout             & \{\textbf{0.1}, 0.2, 0.3\} \\
attention head      & 4 \\
molecule embedding (GROVER) & \{atom, bond, \textbf{both}\} \\
activation function & \{\textbf{PReLU}, ReLU, LeakyReLU, Sigmoid\} \\

\midrule
\textbf{KGE} \\
model                   & \{\textbf{TransE}, RotatE, DistMult, TuckER\}\\
learning rate           & \{1e-3, \textbf{1e-4}, 1e-5, 1e-6\} \\
training epochs         & \{5, 10\} \\
hidden dimension        & \{200, 512, \textbf{1200}\} \\
\midrule
\textbf{K-GNN} \\
GNN model               & \{\textbf{GINE}, GAT, GCN\} \\
$\kappa$-hop            & \{\textbf{2}, 3\}\\
learning rate           & \{1e-3, \textbf{1e-4}, 1e-5, 1e-6\}\\
weight decay            & \{1e-3, 1e-4, \textbf{1e-5}, 1e-6, 1e-7\}\\
hidden dimension      & \{200, 400, 800, \textbf{1200}\} \\
pre-training epochs     & 100\\
edge prediction weight $\lambda_{\textrm{edge}}$   & \{1.0, 1.1, 1.3, \textbf{1.5}, 1.8, 2.0\}\\
node prediction weight $\lambda_{\textrm{node}}$   & \{1.0, 1.1, 1.3, \textbf{1.5}, 1.8, 2.0\}\\
motif prediction weight $\lambda_{\textrm{mot}}$   & \{1.0, 1.1, 1.3, 1.5, \textbf{1.8}, 2.0\}\\

activation function & \{PReLU, ReLU, \textbf{Sigmoid}, \textbf{Softmax}\}\\

\midrule
\textbf{Contrastive Learning} \\
learning rate   & \{1e-4, \textbf{5e-4}, 1e-3, 5e-3\} \\
weight decay    & \{\textbf{1e-3}, 1e-4, 1e-5\} \\
negative sampling ratio ($|\mathcal{D}^{-}|/|\mathcal{D}^{+}|$) & \{4, 8, 16, \textbf{32}, 64\} \\
temperature & \{0.1, 0.3, 0.7, \textbf{1.0}\} \\

\midrule
\textbf{Fine-tuning} \\
batch size              & \{4, 16, \textbf{32}, 64, 128\} \\
inital learning rate (for Noam learning rate scheduler)    & \{1e-3, \textbf{1e5-3}, 1e-2, 1e-1, 1, 10\} \\
maximum learning rate (for Noam learning rate scheduler) & 1e-3 \\
final learning rate (for Noam learning rate scheduler) & 1e-4 \\
warmup epochs   & 2 \\
training epochs    & 20 \\
fold number     & \{4, \textbf{5}, 6\} \\
data splitting & scaffold splitting \\
MLP hidden size & \{100, \textbf{200}, 500\} \\
MLP layer number  &\{1,\textbf{2},3,4\} \\
activation function & \{\textbf{ReLU}, LeakyReLU, PReLU, tanh, SELU\} \\

\bottomrule
\end{tabular}
}
\label{tb:hyperparams}
\end{table*}
\begin{figure}[!h]
\includegraphics[width=\linewidth]{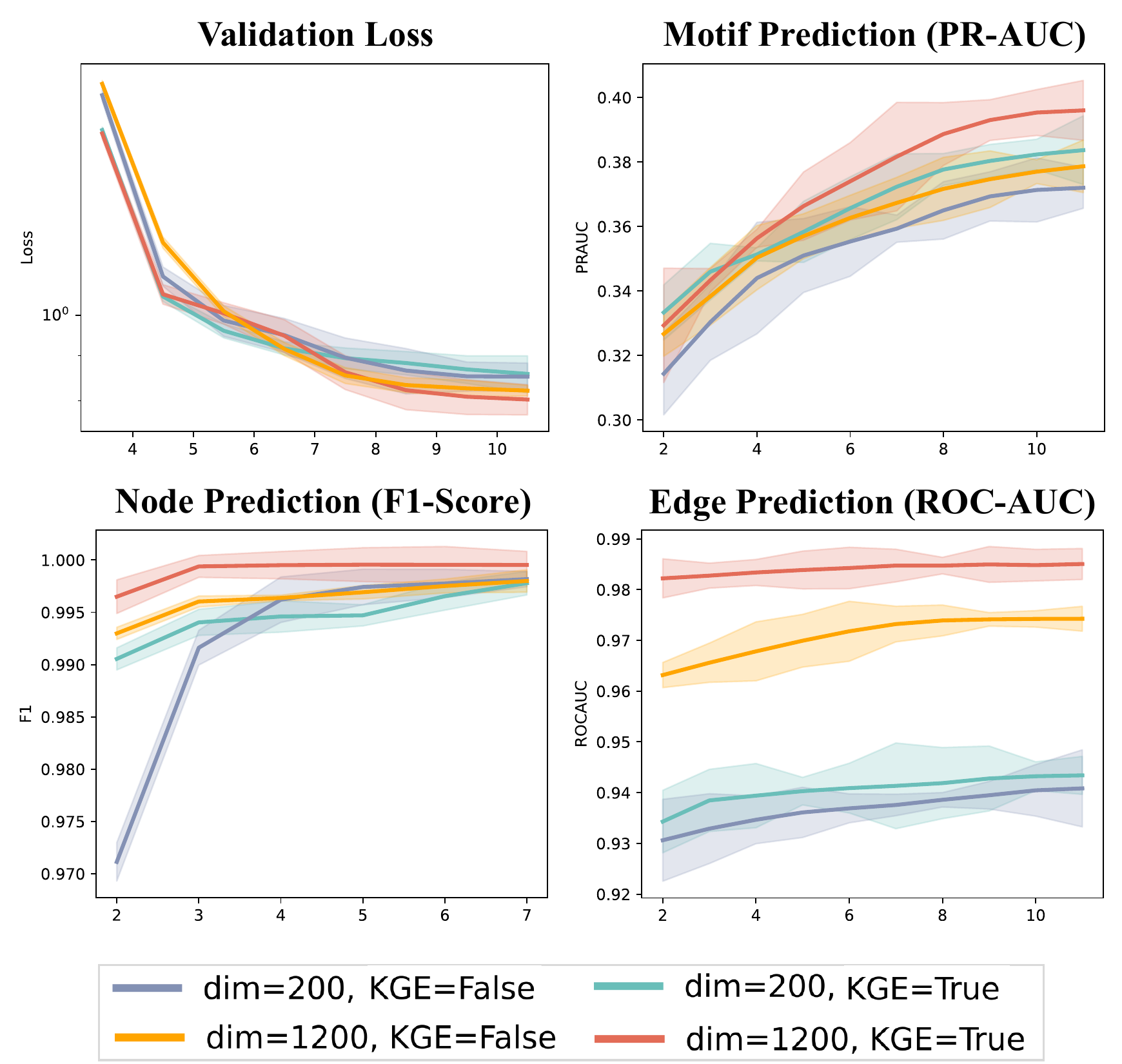}
\caption{\textbf{Performance of knowledge graph-level pre-training tasks.} We report the mean and standard deviation based on five runs with different random seeds.}
\label{fig:kgnn_pretrain}
\end{figure}
\begin{table*}[!htp]
\centering
\caption{Description of Classification Datasets}
\resizebox{0.8\linewidth}{!}{\begin{tabular}{lccp{9cm}}
\toprule
\textbf{Dataset}   &\textbf{\# Molecules} &\textbf{\# Tasks} & \textbf{Description} \\
\midrule

\textbf{BBBP}~\citep{martins2012bayesian}    &2039&1     &The Blood-Brain Barrier Penetration (BBBP) dataset aids drug discovery, especially for neurological disorders. It characterizes a compound's ability to cross the blood-brain barrier, influencing treatment efficacy for brain disorders. \\
\midrule

\textbf{SIDER}~\citep{kuhn2016sider}   &1427&27    &The Side Effect Resource (SIDER) provides adverse effects data of marketed medications. This is crucial for pharmacovigilance, enabling potential side effects predictions of new compounds based on molecular properties. \\
\midrule

\textbf{ClinTox}~\citep{gayvert2016data} &1478&2     &ClinTox compares drugs that gained FDA approval versus those rejected due to toxic concerns. This assists researchers in anticipating toxicological profiles of new compounds. \\
\midrule

\textbf{BACE}~\citep{subramanian2016computational}    &1513&1     &The BACE dataset offers insights into potential inhibitors for human $\beta$-secretase 1 (BACE-1), an enzyme linked to Alzheimer's. It's vital for neurological drug discovery targeting Alzheimer's treatments. \\
\midrule

\textbf{Tox21}~\citep{tox212017}   &7831&12    &Tox21 offers a comprehensive toxicity profile of compounds. Central to the 2014 Tox21 Data Challenge, it aims at enhancing predictions for toxic responses to ensure safer drug design. \\
\midrule

\textbf{ToxCast}~\citep{richard2016toxcast} &8575&617   &ToxCast provides toxicity labels from high-throughput screenings, enabling swift evaluations and guiding early drug development stages. \\

\bottomrule
\end{tabular}
}

\label{tb:downstream_cls}
\end{table*}

\begin{table*}[!h]
\centering
\caption{Description of Regression Datasets}
\resizebox{0.8\linewidth}{!}{\begin{tabular}{lccp{9cm}}
\toprule
\textbf{Dataset}   &\textbf{\# Molecules} &\textbf{\# Tasks} & \textbf{Description} \\
\midrule

\textbf{FreeSolv}~\citep{mobley2014freesolv} &642&1 &A dataset that brings together information on the hydration free energy of molecules in water. The dual presence of experimental data and alchemical free energy calculations offers researchers a robust platform to understand solvation processes and predict such properties for novel molecules.\\
\midrule

\textbf{ESOL}~\citep{delaney2004esol} &1128 &1 &Understanding the solubility of compounds is fundamental in drug formulation and delivery. The ESOL dataset chronicles solubility attributes, providing a structured framework to predict and modify solubility properties in drug design. \\
\midrule

\textbf{Lipophilicity}~\citep{gaulton2012chembl} &4200&1   &Extracted from the ChEMBL database, this dataset focuses on a compound's affinity for lipid bilayers—a key factor in drug absorption and permeability. It provides valuable insights derived from octanol/water distribution coefficient experiments.\\
\midrule

\textbf{QM7}~\citep{blum2009970}   &6830&1   &A curated subset of GDB-13, the QM7 dataset houses details on computed atomization energies of stable, potentially synthesizable organic molecules. It provides an arena for validating quantum mechanical methods against empirical data, bridging computational studies with experimental chemistry. \\
\midrule

\textbf{QM8}~\citep{ramakrishnan2015electronic}  &21786&12    &A more extensive dataset, QM8 encompasses computer-generated quantum mechanical properties. It details aspects like electronic spectra and the excited state energy of molecules, offering a robust resource for computational chemists aiming to predict or understand such attributes. \\
\bottomrule
\end{tabular}
}
\label{tb:downstream_rgr}
\end{table*}
\begin{figure*}[!t]
    \centering
    \includegraphics[width=\linewidth]{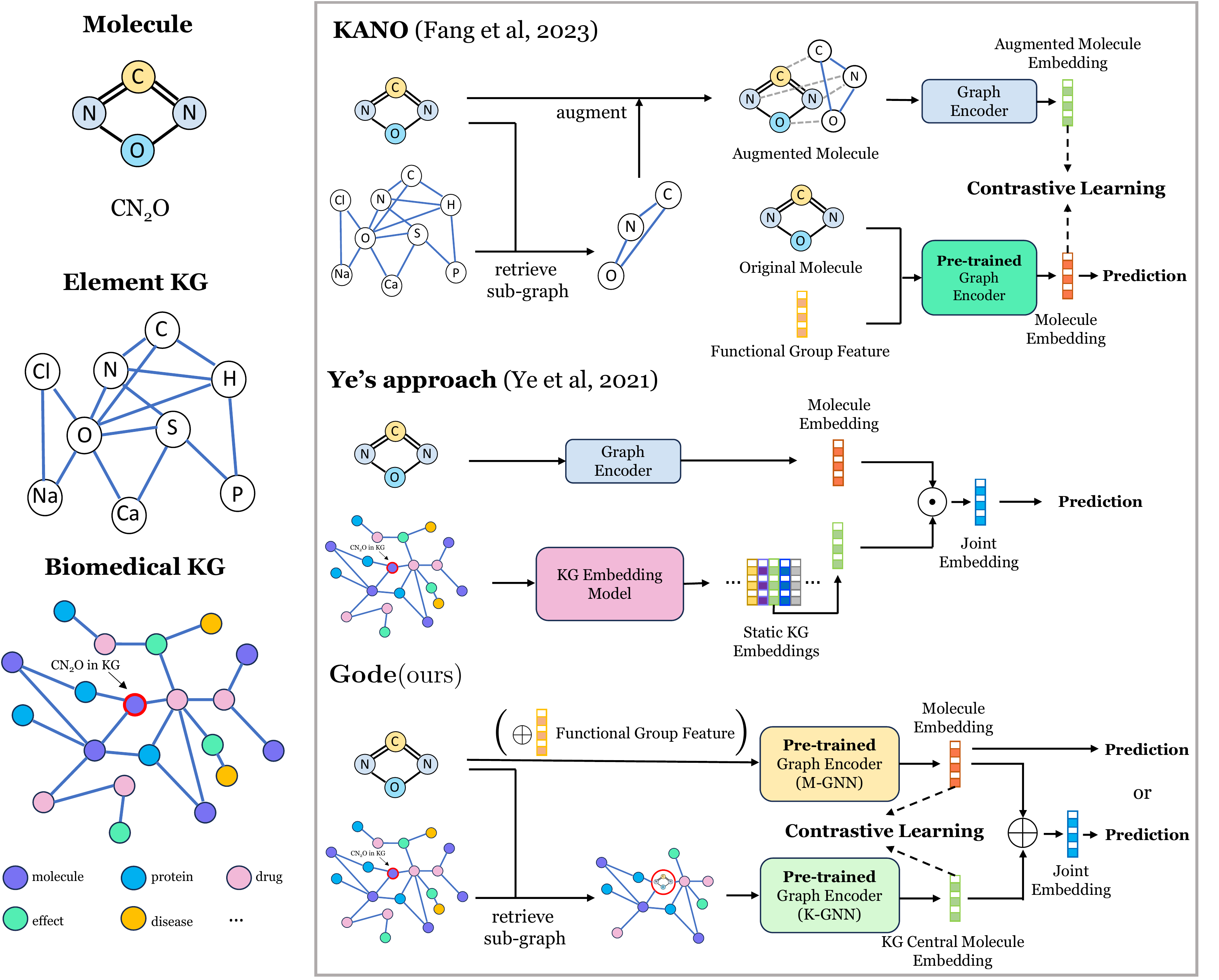}
    \caption{ \textbf{An overview of the difference between \textsc{Gode} with similar works (KGE\_NFM by \citeauthor{ye2021unified} and KANO by \citeauthor{fang2023knowledge}) leveraging both knowledge graph and molecule.} Details such as pre-training strategies or KG embedding initialization are not depicted, for clearer presentations.
    }
    \label{fig:compare}
\end{figure*}
\section{Implementation Details}
In this section, we provide details about the implementation of our proposed method, \textsc{Gode}, as well as the baseline models used for comparison.
\subsection{KG-level Pre-training}

For K-GNN pre-training, we studied how KGE embedding initialization and embedding dimensionality affect the performance of each sub-task in Eq. \ref{eq:kgnn_pretrain_tasks} in the main paper. The representative results are shown in Figure~\ref{fig:kgnn_pretrain}, which illustrate the pivotal role of KGE embedding initialization in augmenting the efficacy of K-GNN pre-training tasks. This advantage manifests as enhanced task performance and consistently diminished validation loss, signifying sharper predictions. The data also indicates a direct relationship between embedding dimensionality and pre-training quality: larger dimensions consistently yield superior results.

\subsection{Hyper-parameter Study of \textsc{Gode}}
We summarize our extensive hyper-parameter study in Table \ref{tb:hyperparams}. Following previous works~\citep{rong2020selfsupervised, fang2023knowledge}, we use RDKit to extract additional features (dimension 200) of M-GNN. 

\subsection{Baseline Models}
In this work, we compare \textsc{Gode} to 13 baseline methods, including GCN~\citep{kipf2016semi}, GIN~\citep{xu2018powerful}, SchNet~\citep{schutt2017schnet}, MPNN~\citep{gilmer2017neural}, DMPNN~\citep{yang2019analyzing}, MGCN~\citep{lu2019molecular}, N-GRAM~\citep{liu2019n}, Hu et al~\citep{hu2019strategies}, GROVER~\citep{rong2020selfsupervised}, MGSSL~\citep{zhang2021motifbased}, KGE\_NFM~\citep{ye2021unified}, MolCLR~\citep{wang2102molclr}, and KANO~\citep{fang2023knowledge}.

Similar as KANO~\citep{fang2023knowledge}\footnote{see ``Baseline experimental setup'' in ``Supplementary information'' on \url{https://www.nature.com/articles/s42256-023-00654-0}.}, we reuse the results of GCN, GIN, SchNet, MGCN, N‐GRAM, and HU et al. (2019) from the paper of MolCLR~\citep{wang2102molclr},
and reuse the results of MGSSL from its original paper. We reuse the results of MPNN, DMPNN, and MolCLR (default setup) from the paper of KANO, which ensures fair comparison in the same setup.
We reproduced GROVER, MolCLR (with the GTransformer~\citep{rong2020selfsupervised} backbone), KGE\_NFM (with our MolKG), and KANO based on the source code they provided\footnote{GROVER: \url{https://github.com/tencent-ailab/grover}}\footnote{MolCLR: \url{https://github.com/yuyangw/MolCLR}}\footnote{KGE\_NFM: \url{https://zenodo.org/records/5500305}}\footnote{KANO: \url{https://github.com/HICAI-ZJU/KANO}}.
Below are the implementation details.

\noindent\textbf{GROVER}~\citep{rong2020selfsupervised}: We use the same implementation setup as the original paper. We use node embeddings from both node-view and edge-view GTransformers with self-attentive READOUT function for fine-tuning and property prediction. The mean value of the prediction scores from two GTransformers is used for prediction.

\noindent $\textbf{MolCLR}_{\textbf{GTrans}}$~\citep{wang2102molclr}: We change the backbone molecule encoder of MolCLR to GTransformer. For a fair comparison, we pre-train node-view and edge-view GTransformers (hidden dimension 1200) separately with MolCLR's contrastive learning framework. For fine-tuning and prediction, we take the same setting as GROVER.

\noindent
\textbf{KGE\_NFM}~\citep{ye2021unified}: We treat this approach as a general framework fusing molecule graph with static KGE embedding (see Appedix \ref{ap:kge_nfm}).  we use node-view and edge-view pre-trained GTransformers ($\text{GROVER}_{\text{Large, GTrans}}$) as the molecule encoders and use DistMult as the static KGE method (hidden dimension 1200). For fine-tuning, we use the original paper's NFM integration and update the node-view and edge-view GTransformers separately. We take the mean of the scores from two models for the property prediction.

\noindent
\textbf{KANO}~\citep{fang2023knowledge}: We implement KANO with two backbone models: CMPNN~\citep{song2020communicative} and GTransformer, where the former is the original paper's implementation, and the latter is ours. For $\text{KANO}_{\text{CMPNN}}$, We keep the same setup described by the original paper and the provided code. For $\text{KANO}_{\text{GTrans}}$, we separately train node-view and edge-view GTransformers with KANO's contrastive-based pre-training strategy and fine-tune the pre-trained encoders with KANO's prompt-enhanced fine-tuning strategy. The mean value of prediction scores is taken for property prediction. 



\section{Datasets of Downstream Tasks}
We introduce the datasets/tasks in Tables \ref{tb:downstream_cls}
and \ref{tb:downstream_rgr}.

\section{Comparison with Similar Studies}
\label{ap:similar_work}
We present a comparative analysis between the proposed \textsc{Gode} framework and two notable prior works in molecular property prediction that similarly leverage knowledge graph integration: KGE\-NFM~\citet{ye2021unified} and KANO~\citet{fang2023knowledge}. Figure~\ref{fig:compare} provides a schematic overview highlighting the key architectural differences and similarities among the methods.

\end{document}